\documentclass[runningheads]{llncs}

\usepackage{eccv}

\usepackage{eccvabbrv}

\usepackage{graphicx}
\usepackage{booktabs}

\usepackage[accsupp]{axessibility}  

\usepackage{hyperref}

\usepackage{orcidlink}

\usepackage{multirow} 
\usepackage{multicol}
\usepackage{subcaption}
\usepackage{graphicx}
\usepackage{booktabs}
\usepackage{colortbl}
\usepackage{booktabs}

\usepackage{pifont}

\usepackage{array}
\newcolumntype{I}{!{\vrule width 3pt}}
\newlength\savedwidth

\newlength\savewidth

\begin{document}

\title{CamoTeacher: Dual-Rotation Consistency Learning for Semi-Supervised Camouflaged Object Detection} 

\titlerunning{CamoTeacher}


\author{
Xunfa Lai\inst{1} \and
Zhiyu Yang\inst{1} \and
Jie Hu\inst{2} \and
Shengchuan Zhang\inst{1}$^{\ast}$\and
Liujuan Cao\inst{1} \and
Guannan Jiang\inst{2} \and
Zhiyu Wang\inst{2} \and
Songan Zhang\inst{3} \and
Rongrong Ji\inst{1} 
}

\renewcommand{\thefootnote}{\fnsymbol{footnote}}
\footnotetext[1]{Corresponding authors}
\authorrunning{X.Lai et al.}


\institute{
Key Laboratory of Multimedia Trusted Perception and Efficient Computing, Ministry of Education of China, Xiamen University \and
Contemporary Amperex Technology Co. Limited \and
Shanghai Jiao Tong University
}

\maketitle

\begin{abstract}
Existing camouflaged object detection~(COD) methods depend heavily on large-scale pixel-level annotations.
However, acquiring such annotations is laborious due to the inherent camouflage characteristics of the objects.
Semi-supervised learning offers a promising solution to this challenge.
Yet, its application in COD is hindered by significant pseudo-label noise, both pixel-level and instance-level.
We introduce CamoTeacher, a novel semi-supervised COD framework, utilizing Dual-Rotation Consistency Learning~(DRCL) to effectively address these noise issues.
Specifically, DRCL minimizes pseudo-label noise by leveraging rotation views' consistency in pixel-level and instance-level.
First, it employs Pixel-wise Consistency Learning~(PCL) to deal with pixel-level noise by reweighting the different parts within the pseudo-label.
Second, Instance-wise Consistency Learning~(ICL) is used to adjust weights for pseudo-labels, which handles instance-level noise.
Extensive experiments on four COD benchmark datasets demonstrate that the proposed CamoTeacher not only achieves state-of-the-art compared with semi-supervised learning methods, but also rivals established fully-supervised learning methods.
Our code will be available soon.


\end{abstract}    
\section{Introduction}
\label{sec:intro}

\begin{figure}[tb]
    \centering
    \includegraphics[width=0.9\columnwidth]{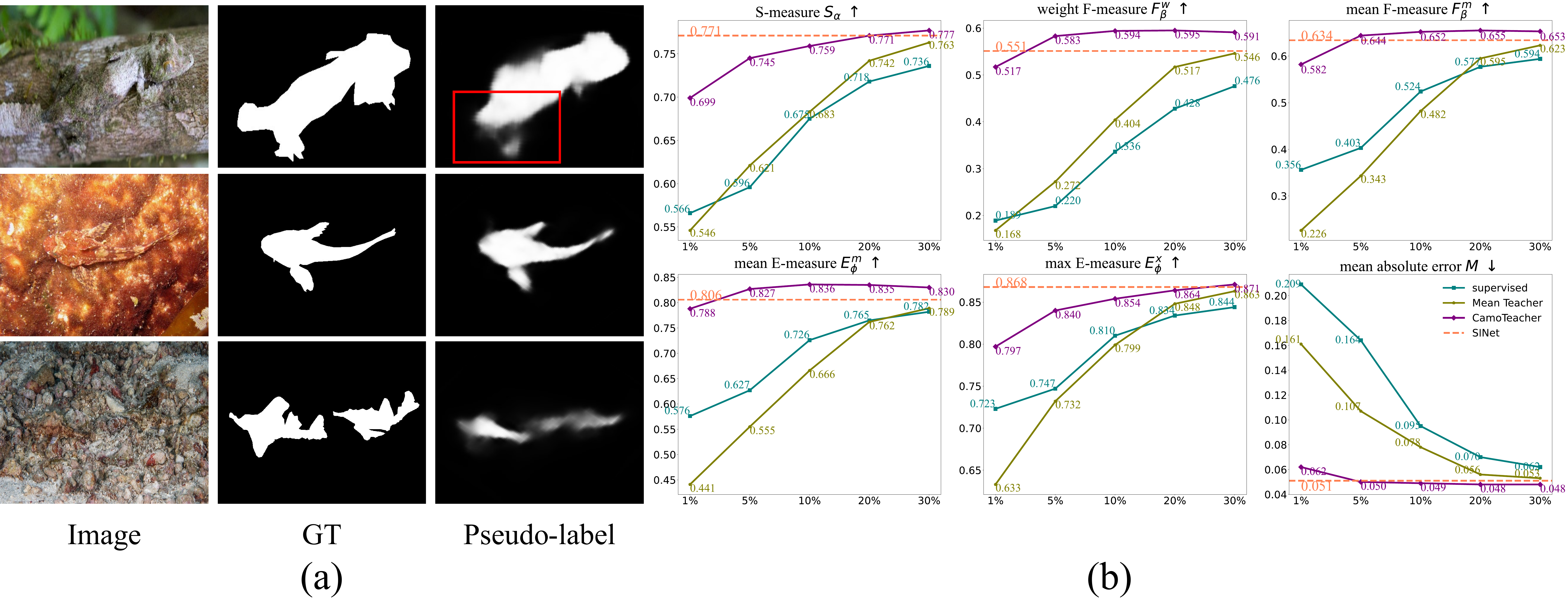}
    \caption{
    (a) Illustration of pixel-level and instance-level noise.
    Pixel-level noise refers to varying degrees of noise in different parts within an instance. 
    Instance-level noise pertains to varying degrees of noise among different instances. 
    (b) Visualization of performance trends on COD10K\cite{COD10K} under different percentages of labeled data.}
    \label{fig:1}
\end{figure}

Camouflaged object detection (COD) aims to identify objects that blend seamlessly into their environments including animals or artificial entities that possess protective coloring and have the ability to blend into their surroundings, a task complicated by low contrast, similar textures, and blurred boundaries~\cite{COD10K, SINetv2, DTINet, FSPNet, SegMaR, FEDER, camoformer}.
Unlike general object detection, COD is challenged by these factors, making detection significantly more difficult.
%
Existing methods in COD heavily depend on extensive pixel-level annotated datasets, the creation of which incurs substantial human effort and cost, thus limiting COD's advancement.

To mitigate that, semi-supervised learning~\cite{SSL2021,AugSeg, Active-Teacher, PAIS, mean-teacher} emerges as a promising approach by leveraging both labeled and unlabeled data, yet its application in COD is not straightforward due to complex backgrounds and subtle object boundaries.
The efficacy of semi-supervised learning in COD is significantly compromised by the presence of substantial noise in pseudo-labels. 
Our investigation into pseudo-label noise reveals two primary types: pixel-level noise, indicating variation within a single pseudo-label, and instance-level noise, showing variation across different pseudo-labels. This distinction is critical, as it informs our approach to refining pseudo-label quality for improved model training.
(1) Pixel-level noise is characterized by inconsistent labeling within various parts of a pseudo-label.
As demonstrated in~\cref{fig:1}(a), in the first row, the tail of the gecko is visually more camouflaged than its head. The pseudo-label generated by the SINet\cite{COD10K} is less accurate in its tail region (highlighted by the red box).
This observation underscores the inappropriateness of uniformly treating all parts within a pseudo-label.
(2) Instance-level noise refers to the variability in noise levels across distinct pseudo-labels.
As shown in \cref{fig:1}(a), the pseudo-labels of the third row are less accurate compared to that in the second row because the camouflaged objects in the third row is more difficult to detect.
These discrepancies indicate that each pseudo-label contributes variably to model training, emphasizing the need for a nuanced approach to integrating pseudo-label information.

\begin{figure}[tb]
    \centering
    \includegraphics[width=0.9\columnwidth]{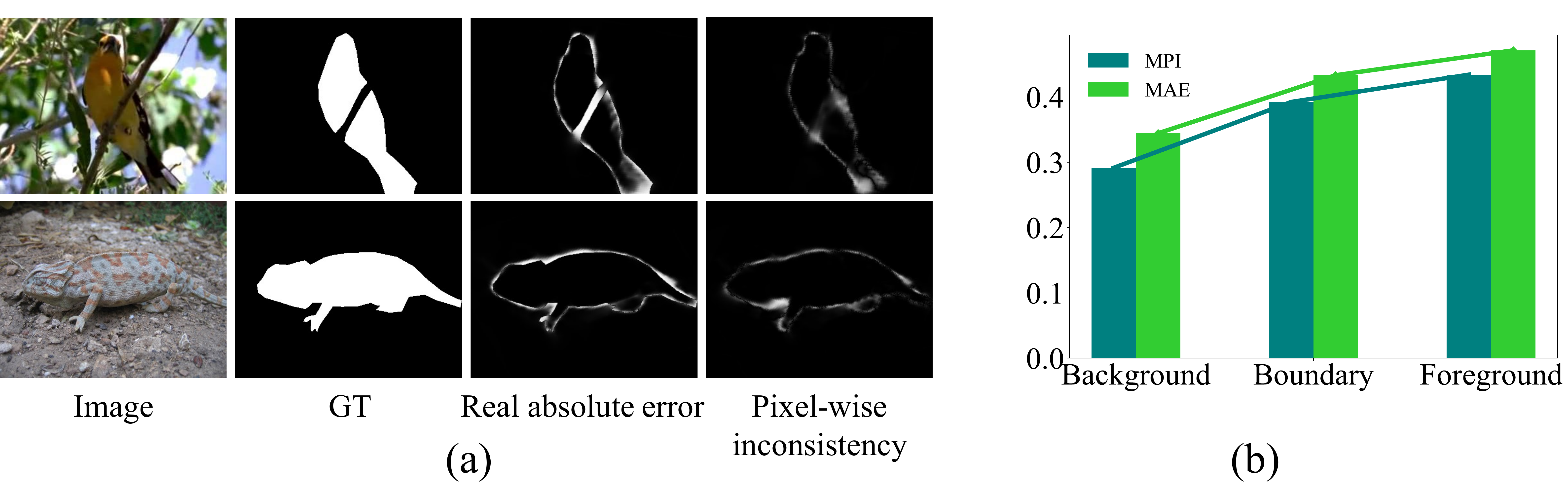}
    \caption{
    (a) The real absolute error is obtained by subtracting the pseudo-label from the ground truth, reflecting the actual noisy level. On the other hand, pixel-wise inconsistency is derived from the subtraction of pseudo-labels from two rotated views. Visually, pixel-wise inconsistency and real absolute error appear very similar.
    (b) We define the area within 20 pixels from the edge as the boundary according to GT. 
    We calculate mean pixel-wise inconsistency (MPI) and mean absolute error (MAE) in background, foreground, and boundary.
    The pseudo-labels are obtained by our proposed model on the test dataset COD10K\cite{COD10K} at a 10\% semi-supervised setting.
    }
    \label{fig:pixel}
\end{figure}

\begin{figure}[tb]
    \centering
    \includegraphics[width=0.9\columnwidth]{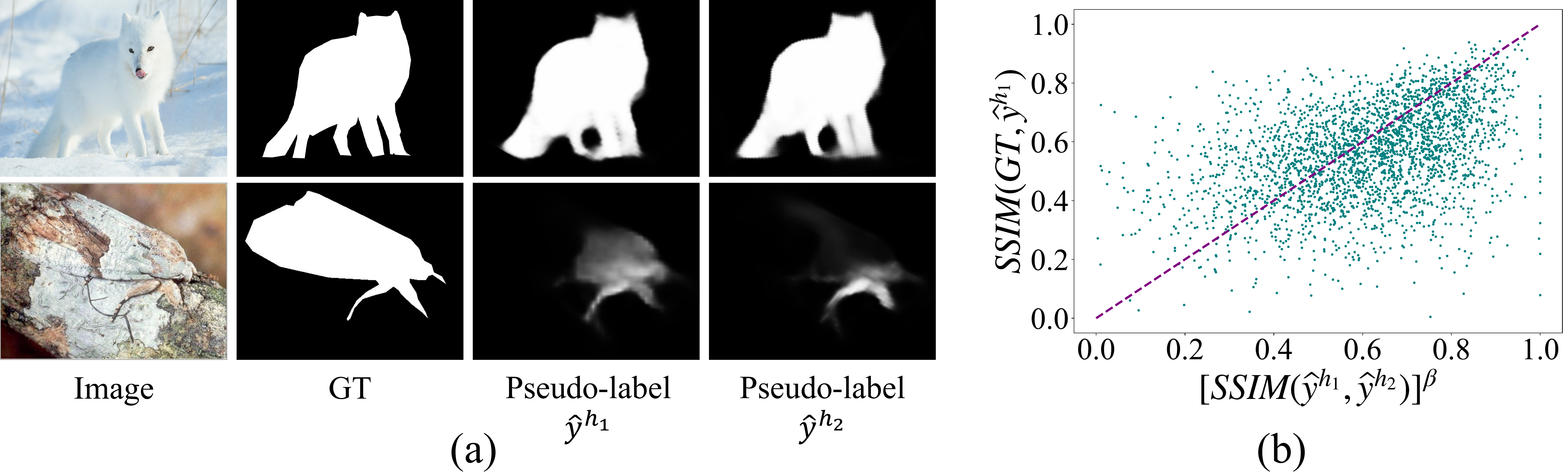}
    \caption{
    (a) The more similar the pseudo-labels are between two rotated views, the lower the noise level and the higher the quality of the pseudo-labels. For example, in the first row, the pseudo-labels are more similar, indicating a lower noise level compared to the pseudo-labels in the second row.
    (b) The positive correlation between instance-wise consistency and the SSIM\cite{SSIM} of pseudo-labels and GT.
    }
    \label{fig:instance}
\end{figure}

To tackle the challenge of evaluating pseudo-label noise without ground truth (GT) for unlabeled data, we introduce two novel strategies based on pixel-wise inconsistency and instance-wise consistency across two rotation views. 
Specifically, for pixel-level noise, we observe that pixel-wise inconsistency—calculated by comparing pseudo-labels from two rotated views—mirrors the actual error relative to GT, as shown in \cref{fig:pixel}(a). 
This relationship, illustrated by the polyline in \cref{fig:pixel}(b), shows a positive correlation between mean pixel-wise inconsistency and the mean absolute error (MAE) across different parts. 
Consequently, areas with higher pixel-wise inconsistency are more prone to inaccuracies, suggesting a need for their de-emphasis during training.
%
%
For instance-level noise, we determine that pseudo-labels with greater similarity across rotation views exhibit lower noise levels, as indicated in \cref{fig:instance}(a). A positive correlation between instance-wise consistency and the SSIM\cite{SSIM} computed by pseudo-labels and GT further supports this observation, depicted in \cref{fig:instance}(b). Thus, pseudo-labels demonstrating higher instance-wise consistency are likely of superior quality and should be prioritized in the learning process.

%
With the insights gained from these observations, we propose a semi-supervised camouflaged object detection framework called CamoTeacher, which incorporates a novel method known as Dual-Rotation Consistency Learning~(DRCL).
Specifically, DRCL operationalizes its strategy through two core components: Pixel-wise Consistency Learning~(PCL) and Instance-wise Consistency Learning~(ICL).
PCL innovatively assigns variable weights to different parts within a pseudo-label, considering the pixel-wise inconsistency across diverse rotation view. 
Simultaneously, ICL adjusts the importance of individual pseudo-labels based on their instance-wise consistency, enabling a nuanced, noise-aware training process.
%
%

We adopt SINet\cite{COD10K} as the base model to implement CamoTeacher and apply it to more classic COD models, \ie CNN-based model SINet-v2 \cite{SINetv2} and SegMaR \cite{SegMaR} and Transformed-based model DTINet \cite{DTINet} and FSPNet \cite{FSPNet}.
Extensive experiments on four COD benchmark datasets,~\ie CAMO \cite{CAMO}, CHAMELEON \cite{CHAMELEON}, COD10K \cite{COD10K}, and NC4K \cite{NC4K}, demonstrate that the proposed CamoTeacher not only achieves state-of-the-art compared with semi-supervised learning methods, but also rivals established fully-supervised learning methods. 
%
Specifically, as shown on \cref{fig:1}(b), with only 20\% labeled data, it nearly reached the performance of fully supervised model on COD10K.

Our contributions can be summarized as follows:
\begin{itemize}
    \item We introduce the first end-to-end semi-supervised camouflaged object detection framework CamoTeacher, offering a simple yet effective baseline for future research in semi-supervised camouflaged object detection.
    
    \item To address the problem of large amounts of noise in pseudo-labels in semi-supervised camouflaged object detection, we proposed the Dual-Rotation Consistency Learning~(DRCL), which includes Pixel-wise Consistency Learning~(PCL) and Instance-wise Consistency Learning~(ICL), allows for the adaptive adjustment of the contribution of pseudo-labels with varying quality, enabling the effective utilization of pseudo-label information.

    \item We conducted extensive experiments on COD benchmark datasets, and achieved significant improvements compared to the fully supervised setting.

\end{itemize}

\section{Related works}
\label{sec:related_works}

\subsection{Camouflaged Object Detection}
Camouflaged Object Detection(COD) is different from traditional visual detection tasks as objects and backgrounds in traditional detection tasks have significant visual distinctions, whereas camouflaged objects are typically concealed within their surrounding environments \cite{COD10K, SINetv2, SegMaR, ZoomNet, DTINet, FSPNet, camoformer}. 
In recent years, many models have been proposed on this task, which can be divided into CNN-based and Transformed-based.
i) CNN-based models: 
SINet\cite{COD10K} is a simple yet effective baseline that views camouflage detection as searching and identification process. 
SINet-v2\cite{SINetv2} significantly improves performance by introducing a neighbor connection decoder and cascaded group-reversal attention design. 
SegMaR\cite{SegMaR} firstly performs initial detection of camouflaged objects and then progressively enlarges the detected regions to achieve precise detection.
FEDER\cite{FEDER} achieves impressive results by conducting feature decomposition and edge reconstruction.
ii) Transformed-based models:
DTINet\cite{DTINet} proposes a dual-task interactive transformer to detect accurate position and detailed boundary.
CamoFormer\cite{camoformer} presents masked separable attention to separate foreground and background regions.
FSPNet\cite{FSPNet} designs a feature shrinkage decoder hierarchically decode locality-enhanced neighboring transformer features.
These outstanding works have achieved excellent results in COD, but they rely on large-scale pixel-level annotations, which are time-consuming and labor-intensive to obtain.
%

 In response to the reliance of full-supervised COD on expensive pixel-level annotations, some unsupervised/weakly-supervised approaches have emerged \cite{UCOS-DA, GenSAM, SCOD, WS-SAM}.
 UCOS-DA \cite{UCOS-DA} formulates the unsupervised camouflaged object segmentation as a source-free unsupervised domain adaptation task and designs a foreground-background-contrastive self-adversarial pipeline.
 SCOD \cite{SCOD} proposes the first weakly-supervised COD method with relabeled scribbles.
 WS-SAM \cite{WS-SAM} uses the provided sparse annotations as prompts for SAM\cite{SAM} to generate segmentation masks to train the model.
 These methods reduce annotation costs but often exhibit lower performance compared to fully-supervised methods. 
 Is it possible to leverage existing annotated data without additional labeling while ensuring accuracy?
 To address this issue, we propose a new benchmark that combines semi-supervised learning with camouflaged object detection.
\subsection{Semi-Supervised Learning}
In traditional supervised learning, training a high-performing model requires a substantial amount of labeled data. 
However, obtaining labeled data in practical applications is often costly and time-consuming. Consequently, the effective utilization of unlabeled data to improve model performance and generalization has become a critical concern. 
Semi-supervised learning addresses this challenge by leveraging unlabeled data to enhance the model's learning process \cite{grandvalet2004semi,lee2013pseudo,oliver2018realistic, zhao2022lassl, sohn2020fixmatch, mean-teacher, Active-Teacher, softmatch, adamatch}, thereby improving its performance and generalization capability. 
Pseudo-labeling \cite{sohn2020simple,wang2021data, softteacher, chen2021semi, wang2022semi, ST++} and consistency regularization \cite{laine2016temporal, lai2021semi, PAIS, PS-MT, UniMatch} are popular techniques in the field of semi-supervised learning. 
These methods have shown great efficacy when combined, producing impressive results.

To address the issue of heavy reliance on labeled data in camouflage object detection, we propose the first end-to-end semi-supervised camouflage object detection method. By integrating semi-supervised learning, our approach aims to enhance the performance of camouflage object detection models while significantly reducing the need for annotated data.
\section{Methodology}
\label{sec:methodology}
%
\subsection{Task Formulation}
Semi-supervised camouflaged object detection aims to leverage a limited amount of annotated data to train a detector capable of identifying objects that seamlessly blend into their surroundings. This task is inherently challenging due to the low contrast between the objects and their background.
Given a camouflaged object detection dataset $D$ for training, we represent the labeled subset with $M$ labeled samples as $D_L =\{x_i^{l}, y_i\}_{i=1}^{M}$ , and the unlabeled subset with $N$ unlabeled samples as $D_U =\{x_i^{u}\}_{i=1}^{N}$ , where $x_i^{l}$ and $x_i^{u}$ represent the input image, $y_i$ represent the corresponding annotation mask of labeled data.
Typically, $D_L$ constitutes a small fraction of the entire dataset $D$, highlighting the semi-supervised learning scenario with $M \ll N$.
%
The emphasis on $M \ll N$ underscores the challenge and opportunity within semi-supervised learning: to enhance detection capabilities by utilizing the untapped potential of the unlabeled data $D_U$, which far exceeds the labeled subset $D_L$. 
%

\subsection{Overall Framework}

\begin{figure*}[tb]
    \centering
    \includegraphics[width=1.00\columnwidth]{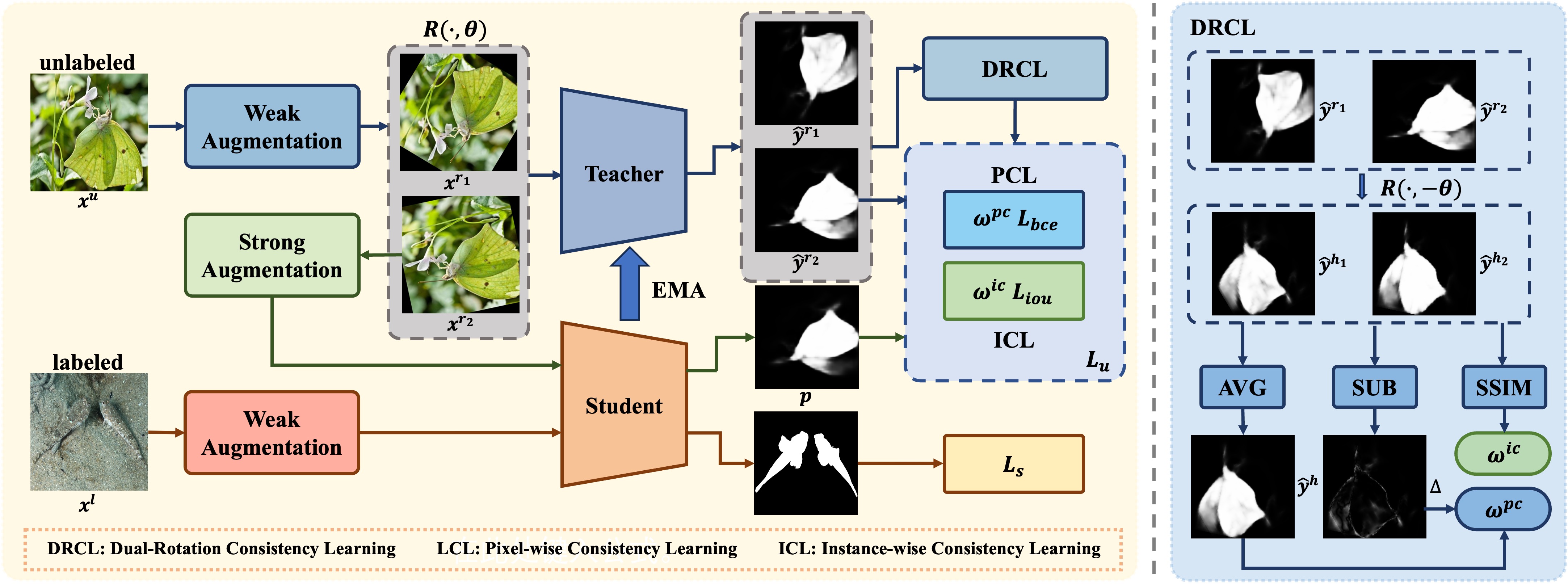}
    \caption{\textbf{The overall pipeline of our CamoTeacher.} CamoTeacher consists of a teacher model and a student model. The teacher model generates pseudo-labels to optimize the student model, while the teacher model is updated through EMA \cite{mean-teacher} from the student. To mitigate pseudo-labels' noise, we introduced Dual-Rotation Consistency Learning (DRCL), which involves Pixel-wise Consistency Learning (PCL) and Instance-wise Consistency Learning (ICL). DRCL weights the loss function based on pixel-wise inconsistency and instance-wise consistency in pseudo-labels across different rotation views.}
    \label{fig:pipeline}
\end{figure*}

As shown in \cref{fig:pipeline}, we adopt Mean Teacher \cite{mean-teacher} as preliminary to achieve end-to-end semi-supervised camouflaged object detection framework, which contains two COD models with the same structure, \ie, teacher model and student model, parameterized by $\Theta_t$ and $\Theta_s$, respectively. The teacher model generates pseudo-labels, which are then used to optimize the student model. The overall loss function $L$ can be defined as:

\begin{equation}
   L = L_s + \lambda_u L_u , 
\end{equation}
where $L_s$ and $L_u$ denote supervised loss and unsupervised loss respectively, $\lambda_u$ is the weight of unsupervised loss to balance the loss terms.
Following the classic COD method, we use binary cross-entropy loss $ L_{bce} $ for training.

During the training process, we employ a combination of weak augmentation $\mathcal{A}^w(\cdot)$ and strong augmentation $\mathcal{A}^s(\cdot)$ strategies. Weak augmentation is applied to labeled data to mitigate overfitting, while unlabeled data undergo various data perturbations under strong augmentation to create diverse perspectives of the same image. The supervised loss $L_s$ is defined by:

\begin{equation}
    L_s = \frac{1}{M} \sum\limits^{M}_{i=1} L_{bce}(F(\mathcal{A}^w(x_i^l);\Theta_s), y_i) ,
\end{equation}
where $F(\mathcal{A}(x_i);\Theta)$  represents the detection results obtained by the model $\Theta$ for the $i$-th image with  augmentations $\mathcal{A}(\cdot)$.
%
For unlabeled images, we first apply weak augmentation, denoted as $\mathcal{A}^w(\cdot)$, before forwarding them to the teacher model. This initial step is crucial for generating reliable pseudo-labels $\widehat{y_i}$ under variations that do not significantly alter the core features of the images. These pseudo-labels serve as a form of soft supervision for the student model. Subsequently, the same images are subjected to strong augmentation, $\mathcal{A}^s(\cdot)$, and then passed to the student model. This process introduces a higher level of variability and complexity, simulating more challenging conditions for the student model to adapt to. The student model generates predictions $p_i$ based on these strongly augmented images, leveraging the pseudo-labels $\widehat{y_i}$ for guidance in learning from the unlabeled data. It can be formulated as:
\begin{equation}
    \widehat{y_i} = F(\mathcal{A}^w(x_i^u);\Theta_t), \ p_i = F(\mathcal{A}^s (\mathcal{A}^w(x_i^u));\Theta_s)  . 
\end{equation}
%
Hence, the unsupervised loss $ L_u $ can be expressed as:
\begin{equation}
    L_u = \frac{1}{N} \sum\limits^{N}_{i=1} L_{bce}(p_i, \widehat{y_i}).
\end{equation}

Finally, the student model is intensively trained using the total loss $L$, which encapsulates both the supervised and unsupervised learning aspects of the semi-supervised framework. This approach ensures that the student model benefits from both labeled and pseudo-labeled data, enhancing its detection capabilities. Concurrently, the teacher model is methodically updated through the Exponential Moving Average~(EMA) mechanism \cite{mean-teacher}, which effectively distills students’ knowledge while preventing noise, formulated as:
 \begin{equation}
    \Theta_t \leftarrow \eta \Theta_t + (1 - \eta)\Theta_s  ,
 \end{equation}
where $\eta$ is hyper-parameters denotes the keeping rate.

\subsection{Dual-Rotation Consistency Learning }



Due to the camouflaging nature of the objects, pseudo-labels contain a substantial amount of noise, and directly using them to optimize the student model potentially harm the model's performance.
To address this problem, one of the most intuitive possible methods is to set a fixed high threshold to filter high-quality pseudo-labels, but this results in a low recall rate and makes it difficult to fully utilize the supervision information of pseudo-labels.
%
%
To this end, we propose Dual-Rotation Consistency Learning (DRCL) to dynamically adjust the weight of the pseudo-labels to reduce the impact of noise.

We randomly apply two separate rotations to the image $ x_i $, which has previously been flipped and random resized, resulting in two distinct rotation views $ x_i^{r_1} $ and $ x_i^{r_2} $.
\begin{equation}
     x_i^{r_1} = R(\mathcal{A}^w(x_i), \theta_1), \ x_i^{r_2} = R(\mathcal{A}^w(x_i), \theta_2),
\end{equation}
where $x_i^{r} = R(x_i, \theta)$  represents rotate the input image $x_i$ by $\theta$ degrees.
%
We input the acquired rotation view into the teacher model to obtain the corresponding prediction, \ie $\widehat y_i^{r} = F(x_i^{r}; \Theta_t)$ . Subsequently, we apply an opposite rotation of $ -\theta $ to return the prediction to its original horizontal orientation, resulting in $\widehat y_i^{h_1}$ and $\widehat y_i^{h_2}$ , facilitating the calculation of prediction inconsistency under different rotation views, respectively,
\begin{equation}
    \widehat y_i^{h_1} = R(\widehat y_i^{r_1}, -\theta_1), \ \widehat y_i^{h_2} = R(\widehat y_i^{r_2}, -\theta_2).
\end{equation}
%
Note that rotation introduces black boundary areas, which are not involved in the implement of DRCL.

Since the noise levels vary across different regions of pseudo-labels and among different pseudo-labels, we introduce PCL and ICL dynamically adapt the contributions of different pixels within pseudo-labels and across various pseudo-labels.

\subsubsection{Pixel-wise Consistency Learning}
%

We subtract the horizontal predictions $ \widehat y_i^{h_1} $ and $\widehat y_i^{h_2}$ at the pixel level to obtain the pixel-wise inconsistency $\Delta_i$,
\begin{equation}
    \Delta_i = | \widehat y_i^{h_1}  - \widehat y_i^{h_2} |.
\end{equation}


The pixel-wise inconsistency $ \Delta_i $ across different views reflects the reliability of pseudo-labels.
However, in cases where both rotation views' prediction are close to 0.5, $ \Delta_i $ cannot effectively distinguish them. 
These predictions exhibit high uncertainty, meaning they cannot be definitively classified as foreground or background, and are likely to represent noisy labels. Consequently, it becomes necessary to diminish their influence by lowering their weights.
Hence, we calculate the average of horizontal predictions $ \widehat y_i^{h} $,
\begin{equation}
    \widehat y_i^{h} = avg ( \widehat y_i^{h_1} , \widehat y_i^{h_2} ),
\end{equation}
where $avg(\cdot, \cdot)$ represents calculate the average of two inputs in pixel-level, and the use its L2 distance from 0.5 as a component to adjust the weights.


As a result, we derive the pixel-wise consistency weight $ \omega_i^{pc} $ based on the pixel-wise inconsistency among different rotation views, as:
\begin{equation}
    \omega_i^{pc} = (1 - \Delta_i^{\alpha})||\widehat y_i^{h} - \mu||_2^2 ,\label{wlc}
\end{equation}
where $\alpha $ is a hyper-parameter and $\mu=0.5$.
This dynamic pixel-wise consistency weight $ \omega_i^{pc} $ assigns greater weight to regions with consistent predictions across various rotation views, while regions with inconsistent predictions receive a smaller weight.

In conclusion, we formulate the PCL loss $ L_u^{PC} $ as: 
\begin{equation}
    \label{unsup_loss}
    \begin{split}
        L_u^{PC} &= \frac{1}{N} \sum\limits^{N}_{i=1} \omega_{i}^{pc} L_{bce}(p_{i}, \widehat {y}_{i}^{r_1}) \\
        &= - \frac{1}{NHW} \sum\limits^{N}_{i=1} \sum\limits^{H \times W}_{j=1} \omega_{i, j}^{pc} [\widehat {y}_{i, j}^{r_1}\log p_{i, j} \\
        & \quad \quad \quad \quad \quad \quad + (1 - \widehat {y}_{i, j}^{r_1})\log (1-p_{i, j})] ,
    \end{split}
\end{equation}
adaptively adjusts the weight of each pixel to ensure comprehensive supervision of student model while avoiding confirmation bias.

\subsubsection{Instance-wise Consistency Learning}

The degree of camouflage varies between different images, leading to significant variations in the quality of pseudo-labels across images. 
Treating all pseudo-labels equally would be unreasonable. 
Unfortunately, for unlabeled images, assessing the quality of pseudo-labels is challenging because there are no GT labels available. 
We have observed a positive correlation between the instance-wise consistency of two rotation views and the quality of pseudo-labels, as quantified by SSIM \cite{SSIM}.
Base on this, we introduce ICL to adjust the contributions of pseudo-labels with varying quality. 
We represent the instance-wise consistency weight $\omega_i^{ic}$ as follows:
\begin{equation}
    \omega_i^{ic} =  (SSIM( \widehat y_i^{h_1} , \widehat y_i^{h_2} ))^{\beta},
\end{equation}
where $\beta$ is a hyper-parameter used to adjust the distribution relationship between instance-wise consistency and pseudo-label quality.

We use intersection-over-union (IoU) loss as the instance-wise restriction, therefore, the ICL loss can be expressed as:
\begin{equation}
    \begin{split}
        L_{u}^{IC} &= \frac{1}{N} \sum\limits^{N}_{i=1} \omega_i^{ic} L_{iou}( p_i , \widehat y_i^{r_1} ) \\
                &= \frac{1}{NHW} \sum\limits^{N}_{i=1} \sum\limits^{H \times W}_{j=1} \omega_i^{ic} \Bigg ( 1 - \frac{ p_{i, j} \widehat {y}_{i,j}^{r_1} }{ p_{i,j} + \widehat {y}_{i, j}^{r_1} - p_{i,j} \widehat y_{i, j}^{r_1} } \Bigg ).
    \end{split} 
\end{equation}

Therefore, the final total $ L $ loss comprises three components: supervised loss $L_s$, PCL loss $L_u^{LC}$, and ICL loss $ L_u^{GC} $ , which can be expressed as:
\begin{equation}
   L = L_s + \lambda_u^{pc} L_u^{PC} + \lambda_{u}^{ic} L_u^{IC},
\end{equation}
where $\lambda_u^{pc}, \lambda_{u}^{ic}$ are the hyper-parameters.

\section{Experiment}
\label{experiment}

\begin{table*}[tb]
\centering


\caption{Experimental results on four COD benchmark datasets includes CAMO, CHAMELEON, COD10K, NC4K. $\dagger$ denotes using the same data augmentation as semi-supervised training. $\uparrow$ indicates the higher the score the better, and $\downarrow$ indicates the lower the better. "n\%" means that "n\%" of the data was utilized as labeled data, with the remainder as unlabeled data. As the data partitioning is random, we conducted 5 fold experiments and the results are the \textbf{average of all 5 folds.} }
\begin{subtable}{\textwidth}
\resizebox{\textwidth}{!}
{
\begin{tabular}{c  | c  c  c  c  c  c   | c  c  c  c  c  c      | c  c  c  c  c  c   }
\toprule
\multicolumn{19}{c}{\textbf{CHAMELEON (76)}} \\ \hline
\multirow{2}{*}{\textbf{Methods}}      & \multicolumn{6}{c|}{\textbf{1\% (41)}}  & \multicolumn{6}{c|}{\textbf{5\% (202)}} & \multicolumn{6}{c}{\textbf{10\% (404)}}\\ 
\cline{2-7} \cline{8-13} \cline{14-19} 
&{$S_{\alpha} \uparrow$} &{$F_{\beta}^w \uparrow$} &{$E_{\phi}^m \uparrow$} &{$E_{\phi}^x \uparrow$} &{$F_{\beta}^m \uparrow$} &{$M \downarrow$}

&{$S_{\alpha} \uparrow$} &{$F_{\beta}^w \uparrow$} &{$E_{\phi}^m \uparrow$} &{$E_{\phi}^x \uparrow$} &{$F_{\beta}^m \uparrow$} &{$M \downarrow$}

&{$S_{\alpha} \uparrow$} &{$F_{\beta}^w \uparrow$} &{$E_{\phi}^m \uparrow$} &{$E_{\phi}^x \uparrow$} &{$F_{\beta}^m \uparrow$} &{$M \downarrow$} \\ \hline

\textbf{Supervised}

&0.605	&0.268 	&0.582 	&0.741 	&0.435    &0.216 
&0.613	&0.283	&0.601	&0.732	&0.439   &0.185
&0.714 	&0.434     &0.725 	&0.829 	&0.592    &0.110   \\ 
\textbf{Supervised$\dagger$} 
&0.635	&0.301	&0.630	&0.766  &0.479	&0.196	
&0.659	&0.331	&0.655	&0.787  &0.499   &0.159
&0.728	&0.454	&0.744	&0.844  &0.607   &0.106\\ 
\textbf{Mean Teacher} 
&0.537     &0.199		&0.418	&0.636	&0.229   &0.204
&0.611    &0.309		&0.524	&0.745	&0.353   &0.137
&0.679	 &0.450		&0.650	&0.812	&0.512  &0.102
\\ 
\textbf{CamoTeacher }
&\textbf{0.652}	&\textbf{0.476} &\textbf{0.714}	&\textbf{0.762} &\textbf{0.558}  &\textbf{0.093}
&\textbf{0.729}	&\textbf{0.587}	&\textbf{0.785}	&\textbf{0.822}	&\textbf{0.656}	&\textbf{0.070}
&\textbf{0.756}	&\textbf{0.617}	&\textbf{0.813}   &\textbf{0.851}	&\textbf{0.684}	&\textbf{0.065}

\\ 

\hline
\end{tabular}
}
\end{subtable}


\begin{subtable}{\textwidth}
\resizebox{\textwidth}{!}{
\begin{tabular}{c  | c  c  c  c  c  c   | c  c  c  c  c  c      | c  c  c  c  c  c   }
\toprule
\multicolumn{19}{c}{\textbf{CAMO (250)}} \\ \hline
\multirow{2}{*}{\textbf{Methods}}      & \multicolumn{6}{c|}{\textbf{1\% (41)}}  & \multicolumn{6}{c|}{\textbf{5\% (202)}} & \multicolumn{6}{c}{\textbf{10\% (404)}}\\ 
\cline{2-7} \cline{8-13} \cline{14-19} 
&{$S_{\alpha} \uparrow$} &{$F_{\beta}^w \uparrow$} &{$E_{\phi}^m \uparrow$} &{$E_{\phi}^x \uparrow$} &{$F_{\beta}^m \uparrow$} &{$M \downarrow$}

&{$S_{\alpha} \uparrow$} &{$F_{\beta}^w \uparrow$} &{$E_{\phi}^m \uparrow$} &{$E_{\phi}^x \uparrow$} &{$F_{\beta}^m \uparrow$} &{$M \downarrow$}

&{$S_{\alpha} \uparrow$} &{$F_{\beta}^w \uparrow$} &{$E_{\phi}^m \uparrow$} &{$E_{\phi}^x \uparrow$} &{$F_{\beta}^m \uparrow$} &{$M \downarrow$}
 \\ \hline

\textbf{Supervised  }
 &0.573 	 &0.271 	 	 	 &0.541 	 &0.704 	 &0.402  &0.251
 &0.592 	 &0.298 			 &0.581 	 &0.704 	 &0.432  &0.216 
 &0.652 	 &0.390 	 		     &0.647 	 &0.760 	 &0.525 &0.159
 \\ 
\textbf{Supervised $\dagger$ } 
 &0.596 	 &0.298 	 	     &0.580       &0.717 	 &0.442 &0.235
 &0.610 	 &0.316 		 	 &0.595 	 &0.720 	     &0.450  &0.205 
 &0.667 	 &0.409 		 	 &0.666 	 &0.779 	 &0.546  &0.153 

 \\ 
\textbf{Mean Teacher } 
 &0.518 	 &0.207 	 	 	 &0.399 	 &0.620 	 &0.227  &0.226
 &0.575 	 &0.286 	 		 &0.482 	 &0.708 	 &0.322  &0.184
 &0.625 	 &0.397 	 		 &0.578 	 &0.773 	 &0.454  &0.150
\\ 
\textbf{CamoTeacher }
&\textbf{0.621}	&\textbf{0.456}	&\textbf{0.669}	&\textbf{0.736}	&\textbf{0.545}   &\textbf{0.136}
&\textbf{0.669}	&\textbf{0.523}	&\textbf{0.711}	&\textbf{0.775}	&\textbf{0.601}   &\textbf{0.122}
&\textbf{0.701}	&\textbf{0.560}		&\textbf{0.742}	&\textbf{0.795}	&\textbf{0.635}  &\textbf{0.112}
 \\ 

\hline
\end{tabular}
}
\end{subtable}


\begin{subtable}{\textwidth}
\resizebox{\textwidth}{!}{
\begin{tabular}{c  |  c  c  c  c  c  c      | c  c  c  c  c  c      | c  c  c  c  c  c   }

\toprule

\multicolumn{19}{c}{\textbf{COD10K (2026)}} \\ \hline
\multirow{2}{*}{\textbf{Methods}}      & \multicolumn{6}{c|}{\textbf{1\% (41)}}  & \multicolumn{6}{c|}{\textbf{5\% (202)}} & \multicolumn{6}{c}{\textbf{10\% (404)}}\\ 
\cline{2-7} \cline{8-13} \cline{14-19} 
&{$S_{\alpha} \uparrow$} &{$F_{\beta}^w \uparrow$} &{$E_{\phi}^m \uparrow$} &{$E_{\phi}^x \uparrow$} &{$F_{\beta}^m \uparrow$} &{$M \downarrow$}

&{$S_{\alpha} \uparrow$} &{$F_{\beta}^w \uparrow$} &{$E_{\phi}^m \uparrow$} &{$E_{\phi}^x \uparrow$} &{$F_{\beta}^m \uparrow$} &{$M \downarrow$}

&{$S_{\alpha} \uparrow$} &{$F_{\beta}^w \uparrow$} &{$E_{\phi}^m \uparrow$} &{$E_{\phi}^x \uparrow$} &{$F_{\beta}^m \uparrow$} &{$M \downarrow$}
 \\ \hline

\textbf{Supervised  }
 &0.566 	 &0.189 	 	 &0.576 	 &0.723 	 &0.356  &0.209 
 &0.596 	 &0.220 	 	 &0.627 	 &0.747 	 &0.403  &0.164
 &0.675 	 &0.336 		 &0.726 	 &0.810 	 &0.524  &0.095 

\\ 
\textbf{Supervised $\dagger$ } 
 &0.582 	 &0.206 	 	 	 &0.609 	 &0.743 	 &0.386  &0.194
 &0.621 	 &0.246 	 	 	 &0.659 	 &0.774 	 &0.444  &0.147
 &0.680 	 &0.338 	 		 &0.726 	 &0.812 	 &0.528  &0.097

 \\ 
\textbf{Mean Teacher } 
 &0.546 	 &0.168   	     &0.441 	 &0.633 	 &0.226   &0.161
 &0.621 	 &0.272 		 &0.555 	 &0.732 	 &0.343  &0.107 
 &0.683 	 &0.404 	 	 &0.666 	 &0.799 	 &0.482   &0.078
 \\ 
\textbf{CamoTeacher }
&\textbf{0.699}	&\textbf{0.517}	&\textbf{0.788}	&\textbf{0.797}	&\textbf{0.582} &\textbf{0.062}
&\textbf{0.745}   &\textbf{0.583}	&\textbf{0.827}	&\textbf{0.840}	&\textbf{0.644} &\textbf{0.050}
&\textbf{0.759}   &\textbf{0.594}	&\textbf{0.836}	&\textbf{0.854}	&\textbf{0.652}	&\textbf{0.049}
 \\ 

\hline

\end{tabular}
}
\end{subtable}

\begin{subtable}{\textwidth}
\resizebox{\textwidth}{!}{
\begin{tabular}{c  | c  c  c  c  c  c   | c  c  c  c  c  c      | c  c  c  c  c  c   }

\toprule

\multicolumn{19}{c}{\textbf{NC4K (4121)}} \\ \hline
\multirow{2}{*}{\textbf{Methods}}      & \multicolumn{6}{c|}{\textbf{1\% (41)}}  & \multicolumn{6}{c|}{\textbf{5\% (202)}} & \multicolumn{6}{c}{\textbf{10\% (404)}}\\ 
\cline{2-7} \cline{8-13} \cline{14-19} 
&{$S_{\alpha} \uparrow$} &{$F_{\beta}^w \uparrow$} &{$E_{\phi}^m \uparrow$} &{$E_{\phi}^x \uparrow$} &{$F_{\beta}^m \uparrow$} &{$M \downarrow$}

&{$S_{\alpha} \uparrow$} &{$F_{\beta}^w \uparrow$} &{$E_{\phi}^m \uparrow$} &{$E_{\phi}^x \uparrow$} &{$F_{\beta}^m \uparrow$} &{$M \downarrow$}

&{$S_{\alpha} \uparrow$} &{$F_{\beta}^w \uparrow$} &{$E_{\phi}^m \uparrow$} &{$E_{\phi}^x \uparrow$} &{$F_{\beta}^m \uparrow$} &{$M \downarrow$}
 \\ \hline

\textbf{Supervised  }
 &0.618 	 &0.291  	 &0.604 	 &0.752 	 &0.474	 &0.213
 &0.649 	 &0.332    &0.660 	 &0.773 	 &0.524 	 &0.179
 &0.727 	 &0.462	 &0.754 	 &0.833 	 &0.643  	 &0.116 

 \\ 
\textbf{Supervised $\dagger$ } 
 &0.638 	 &0.313 	 	 	 &0.639 	 &0.768 	 &0.510  &0.201
 &0.676 	 &0.362 			 &0.689 	 &0.798 	 &0.562  &0.163 
 &0.735 	 &0.471 	 		 &0.760 	 &0.837 	 &0.650  &0.114
\\ 
\textbf{Mean Teacher } 
 &0.541 	 &0.213 		 &0.424 	 &0.637 	 &0.258   &0.193 
 &0.634 	 &0.355 		 &0.556 	 &0.767 	 &0.420 	 &0.140 
 &0.700 	 &0.492 		 &0.670 	 &0.827 	 &0.565	 &0.109 
\\ 
\textbf{CamoTeacher }
&\textbf{0.718}	&\textbf{0.599}	&\textbf{0.779}	&\textbf{0.814}	&\textbf{0.675} &\textbf{0.090}	
&\textbf{0.777}   &\textbf{0.677}	&\textbf{0.834}	&\textbf{0.859}	&\textbf{0.739}	&\textbf{0.071}
&\textbf{0.791}	&\textbf{0.687} &\textbf{0.842}	&\textbf{0.868}	&\textbf{0.746}	&\textbf{0.068}
\\ 

\bottomrule 

\end{tabular}
}
\end{subtable}

\label{tab:result}
\end{table*}		

\begin{table}[tb]
\centering

\caption{Results of CamoTeacher on different COD models with 10\% labeled data on COD10K.}

\resizebox{\linewidth}{!}
{
\begin{tabular}{c  c  | c  c  c c c c}
\toprule
\textbf{Model}  &\textbf{Setting}
&{$S_{\alpha} \uparrow$} &{$F_{\beta}^w \uparrow$} &{$E_{\phi}^m \uparrow$} &{$E_{\phi}^x \uparrow$} &{$F_{\beta}^m \uparrow$} &{$M \downarrow$} \\ \hline
CNN-Based Models \\ \hline

\multirow{2}{*}{SINet-v2 \cite{SINetv2} (TPAMI2022)}
& supervised             & 0.732	& 0.544  & 0.819	& 0.835	& 0.586  & 0.056  \\
& CamoTeacher            &\textbf{0.760}	&\textbf{0.605}	&\textbf{0.835}	&\textbf{0.843}	&\textbf{0.652}  &\textbf{0.051}	  \\ \hline

\multirow{2}{*}{SegMaR \cite{SegMaR} (CVPR2022)}
& supervised             &0.755	 &0.583	 &0.837	&0.851	&0.623  &0.050 \\
    & CamoTeacher &\textbf{0.775} &\textbf{0.629}	&\textbf{0.856}	&\textbf{0.864} &\textbf{0.675}  &\textbf{0.045} \\ \hline

Transformer-Based Models \\ \hline

\multirow{2}{*}{DTINet \cite{DTINet} (ICPR2022)}
& supervised              & 0.771    &0.626           & 0.857      & 0.865         & 0.670        & 0.043   \\
& CamoTeacher              &\textbf{0.794}	&\textbf{0.661}	&\textbf{0.873}	&\textbf{0.882}	&\textbf{0.695}  &\textbf{0.040} \\ \hline

\multirow{2}{*}{FSPNet \cite{FSPNet} (CVPR2023)}
& supervised              &0.789      &0.622           & 0.823                    & 0.867         &0.675         & 0.042   \\
& CamoTeacher              & \textbf{0.817}     & \textbf{0.684}          & \textbf{ 0.884 }                  &  \textbf{0.903 }       & \textbf{0.720 }       & \textbf{0.037}   \\
                
\bottomrule 

\end{tabular}
}
\label{tab:diff_COD_model}
\end{table}	
\begin{table}[tbp]
\centering
\caption{Result of CamoTeacher on COD10K with additional data. We use 100\% train dataset as labeled data and test dataset as the unlabeled data.
"\textsl{U}" represents the unsupervised setting, "\textsl{W}" denotes the weakly-supervised setting, where weakly-supervision results from scribble annotations. "\textsl{L100\%}" indicates using 100\% of the training set as labeled data, "\textsl{U(CAMO)}" denotes utilizing CAMO as unlabeled data, and likewise for others.
}
\renewcommand\arraystretch{1.2}

\begin{subtable}{\linewidth}
\resizebox{\linewidth}{!}{
\begin{tabular}{c c  | c  c  c c c c}
\toprule
\textbf{Model}  &\textbf{Setting}
&{$S_{\alpha} \uparrow$} &{$F_{\beta}^w \uparrow$} &{$E_{\phi}^m \uparrow$} &{$E_{\phi}^x \uparrow$} &{$F_{\beta}^m \uparrow$} &{$M \downarrow$} \\ \hline

Unsupervised \\ \hline
SelfMask \cite{SelfMask} (CVPRW2022) &\textsl{U}    &0.645  &0.440  &0.687  &0.728  &0.478  &0.125  \\ 
FOUND \cite{FOUND} (CVPR2023) &\textsl{U}    &0.670  &0.482  &0.751  &0.753  &0.520  &0.085  \\ 
UCOS-DA \cite{UCOS-DA} (ICCVW2023) &\textsl{U} &0.689  &0.513  &0.740  &0.741  &0.546  & 0.086 \\ 
\hline  

Weakly-Supervised \\ \hline
SCWSSOD \cite{SCWSSOD} (AAAI2021) &\textsl{W}  &0.710  &0.546  &0.805  &-  &-  &0.055 \\
SCOD \cite{SCOD}(AAAI2023)  &\textsl{W}  &0.733 &0.576  &0.832 &0.845   &0.633  &0.049 \\
WS-SAM \cite{WS-SAM} (NeurIPS2023) &\textsl{W}  &0.803  &-  &0.878  &-  &-  &0.038 \\

 \hline

Semi-Supervised \\ \hline
\multirow{4}{*}{ CamoTeacher \tiny{\color{gray}\textit{SINet}}} &\textsl{L100\%}  &0.771    &0.551	&0.806	&0.868	&0.634 &0.051	   \\ 
    & \textsl{L100\%+U(CAMO)} &0.779	&0.554  &0.799	&0.875 &0.632 &0.051\\
 &\textsl{L100\%+U(NC4K)}  &0.785	&0.574	&0.812	&0.879	&0.642  &0.049  \\ 
    &\textsl{L100\%+U(CAMO,NC4K)} &0.787	&0.576  &0.813	&0.881 &0.644 &0.048\\
\bottomrule 
\end{tabular}
}
\end{subtable}

\label{tab:addition_data}
\end{table}	
\begin{table}[t]
\centering
\begin{minipage}[t]{0.49\textwidth}
\centering
    \makeatletter\def\@captype{table}
    \caption{The ablation study of Pixel-wise Consistency Learning (PCL) and Instance-wise Consistency Learning (ICL) with 5\% labeled data on COD10K.}
    \resizebox{0.9\linewidth}{!}{
        \begin{tabular}{ c c  | c  c | c  c  c  c c c}
        \toprule
        \multicolumn{2}{c|}{\textbf{PCL}}    & \multicolumn{2}{c|}{\textbf{ICL}} 
        & \multirow{2}{*}{$S_{\alpha} \uparrow$} 
        & \multirow{2}{*}{$F_{\beta}^w \uparrow$} 
        & \multirow{2}{*}{$E_{\phi}^m \uparrow$} 
        & \multirow{2}{*}{$E_{\phi}^x \uparrow$} 
        & \multirow{2}{*}{$F_{\beta}^m \uparrow$} 
        & \multirow{2}{*}{$M \downarrow$} \\
        \cline{1-2} \cline{3-4}
        \textbf{$L_{bce}$} &\textbf{    $\omega^{pc}$}  & \textbf{$L_{iou}$} & \textbf{    $ \omega^{ic}$} & & & & & & \\ \hline
        \ding{51}       &                &                  &           &0.682	&0.370	&0.636	&0.787	&0.461 &0.091	\\ 
                        &                & \ding{51}        &        &0.734	&0.572		&0.817	&0.821	&0.626   &0.057	         \\
        \ding{51}       &  \ding{51}     &                  &             &0.724	&0.501	&0.738	&0.832	&0.574   &0.058	               \\ 
                        &                & \ding{51}        &  \ding{51}          &0.742	&0.585	&0.828	&0.834	&0.639    &0.052   \\
        \ding{51}       &  \ding{51}     & \ding{51}        &           &0.745	&0.586		&0.830	&0.840	&0.643  &0.051\\
        \rowcolor{gray!20} \ding{51}       &  \ding{51}     & \ding{51}        &  \ding{51}     &\textbf{0.746}	 & \textbf{0.586}		&\textbf{0.830}	&\textbf{0.843}	&\textbf{0.645}   &\textbf{0.050 }         \\ 
                        
        \bottomrule 
        
        \end{tabular}
    }
    \label{tab:ablation_mian}
\end{minipage}
\begin{minipage}[t]{0.49\textwidth}
\centering
    \makeatletter\def\@captype{table}
    \caption{The ablation study of various weight approach of Pixel-wise Consistency Learning (PCL) with 5\% labeled data on COD10K. In this experiment $\alpha$ is set to 1/4.}
    \resizebox{\linewidth}{!}{
        \begin{tabular}{l   | c  c  c c c c}
        \toprule
        \textbf{weight approach} 
        &{$S_{\alpha} \uparrow$} &{$F_{\beta}^w \uparrow$} &{$E_{\phi}^m \uparrow$} &{$E_{\phi}^x \uparrow$} &{$F_{\beta}^m \uparrow$} &{$M \downarrow$} \\ \hline
        $\omega^{pc} = \widehat y^{r} $             &0.698	&0.393	&0.684	&0.804	&0.500 	&0.110   \\  
        $\omega^{pc} = ||\widehat y^{r}-\mu||_2^2 $  &0.717	&0.485	&0.734	&0.822	&0.559 	&0.061\\ 
        
        $\omega^{pc} = 1 - \Delta^{\alpha} $  &0.719	&0.481  &0.720	&0.827	&0.553	&0.062	 \\
        $\omega^{pc} = (1 - \Delta^{\alpha}) \widehat y^{r} $  &0.695	&0.405	&0.696	&0.795	&0.506 	&0.114\\
        \rowcolor{gray!20} $\omega^{pc} = (1 - \Delta^{\alpha}) ||\widehat y^{r}-\mu||_2^2 $   &\textbf{0.724}	&\textbf{0.501}	&\textbf{0.738}	&\textbf{0.832}	&\textbf{0.574}   &\textbf{0.058} \\
        
        \bottomrule 
        
        \end{tabular}
    }
    \label{tab:weight}
\end{minipage}
\end{table}	

\begin{table}[tb]
\centering
\begin{minipage}[t]{0.49\textwidth}
\centering
    \makeatletter\def\@captype{table}
    \caption{The ablation study of hyper-parameters $\alpha$ in Pixel-wise Consistency Learning (PCL) with 5\% labeled data on COD10K. In this experiment $\beta$ is set to 4, $\lambda^{PC}_u$ is set to 8, and $\lambda^{IC}_u$ is set to 0.3.}
    \scalebox{0.9}{
        \begin{tabular}{c|cccccc}
        \toprule
        {\textbf{$\alpha$}}
        &{$S_{\alpha} \uparrow$} &{$F_{\beta}^w \uparrow$} &{$E_{\phi}^m \uparrow$} &{$E_{\phi}^x \uparrow$} &{$F_{\beta}^m \uparrow$} &{$M \downarrow$} \\ \hline
        2               &0.737	&0.565	&0.813	&0.836	&0.629   &0.052  \\ 
        1              &0.739	&0.566	&0.815	&0.832	&0.627 &0.056\\ 
        1/2             &0.745	&0.575	&0.820	&0.839	&0.637   &\textbf{0.051}	  \\ 
        \rowcolor{gray!20} \textbf{1/4}    & \textbf{0.747}	&\textbf{0.579}		        &\textbf{0.825}	&\textbf{0.842}	&\textbf{0.638}  &\textbf{0.051}  \\ 
        1/8             &0.745	&0.578	&\textbf{0.825}	&0.838	&0.636  &0.052	    \\
                
        \bottomrule 
        \end{tabular}
    }
    \label{tab:hyper_alpha}
\end{minipage}
\begin{minipage}[t]{0.49\textwidth}
\centering
    \makeatletter\def\@captype{table}
    \caption{The ablation study of hyper-parameters $\beta$ in Instance-wise Consistency Learning (ICL) with 5\% labeled data on COD10K. In this experiment $\alpha$ is set to 1/4, $\lambda^{PC}_u$ is set to 8, and $\lambda^{IC}_u$ is set to 0.3.}
    \scalebox{0.9}{ 
        \begin{tabular}{c   | c  c  c c c c}
        \toprule
        
        {\textbf{$\beta$}}
        &{$S_{\alpha} \uparrow$} &{$F_{\beta}^w \uparrow$} &{$E_{\phi}^m \uparrow$} &{$E_{\phi}^x \uparrow$} &{$F_{\beta}^m \uparrow$} &{$M \downarrow$} \\ \hline
        1	&0.743	        &0.576	      &0.824	&0.839	&0.636   &\textbf{0.051}  \\
        2	&0.742	        &0.573	      &0.823	&0.837	&0.632	&0.052  \\
        \rowcolor{gray!20} \textbf{4}	&\textbf{0.747}	&\textbf{0.579}	&\textbf{0.825}	&\textbf{0.842}	&0.638 &\textbf{0.051}	 \\
        5	&0.744	        &0.578	      &0.824	&0.841	&\textbf{0.639}	&\textbf{0.051}  \\
        6	&0.745	        &0.575	      &0.822	&0.839	&0.635	&0.052	 \\
                        
        \bottomrule 
        \end{tabular}
    }
    \label{tab:hyper_beta}
\end{minipage}
\end{table}	
\begin{figure}[t]
    \centering
    \includegraphics[width=0.8\columnwidth]{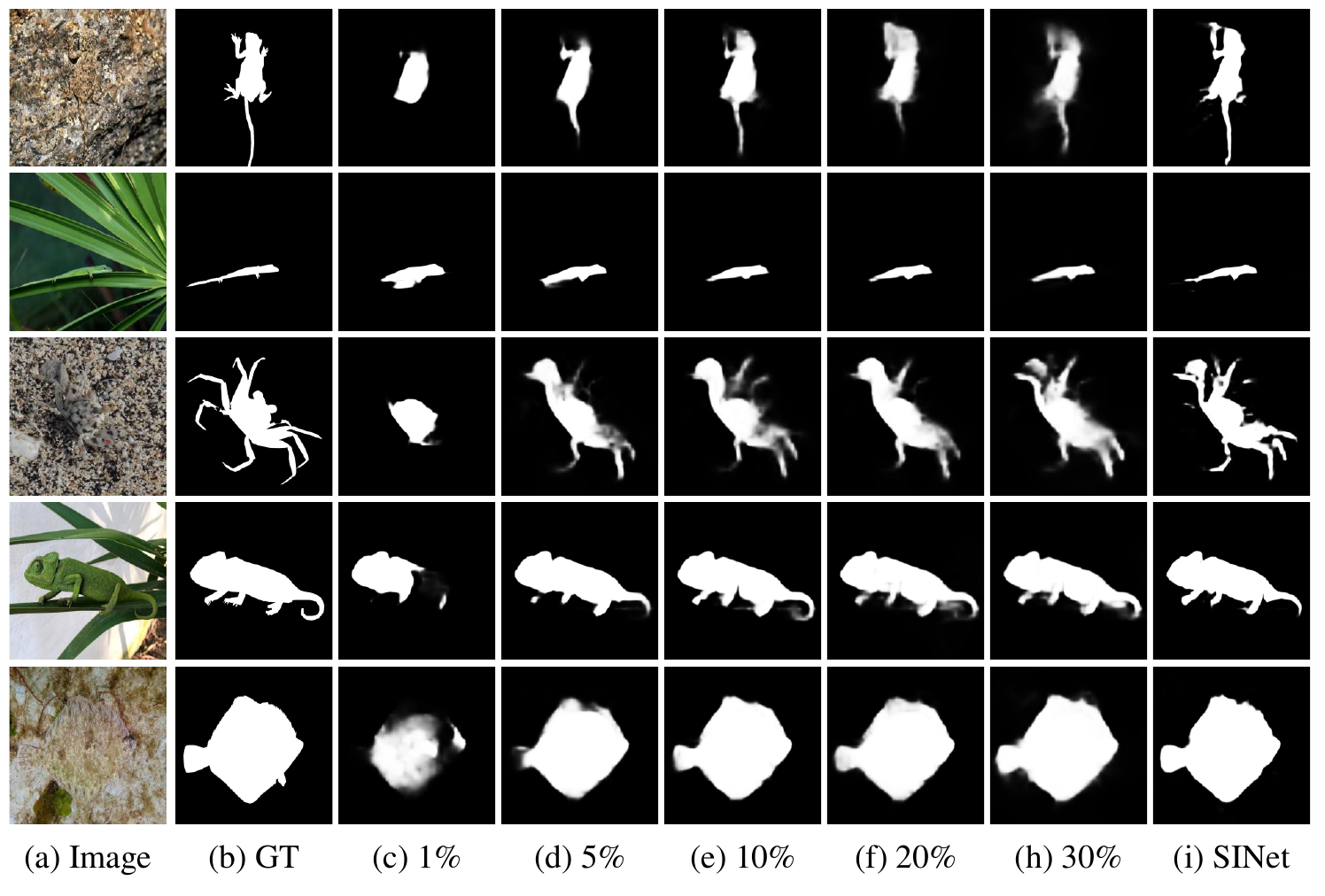}
    \caption{Visualization of predictions in different proportions of labeled data.}
    \label{fig:vis_main}
\end{figure}

\subsection{Experiment Settings}

\noindent\textbf{Dataset.}
We evaluate our CamoTeacher on four benchmark datasets, CAMO \cite{CAMO}, CHAMELEON \cite{CHAMELEON}, COD10K \cite{COD10K}, and NC4K \cite{NC4K}. In CAMO, there are 2500 images, consisting of 1250 camouflage images and 1250 non-camouflage images. CHAMELEON contains 76 manually annotated images. COD10K dataset consists of 5066 camouflage images, 3000 background images, and 1934 non-camouflage images. NC4K is another large-scale COD test dataset that comprises 4,121 images. Based on the data partitioning in previous works \cite{COD10K,ZoomNet,zhang2022preynet}, we used 3040 images from COD10K and 1000 images from CAMO as the training set for our experiments. The remaining images from both datasets were used as the test set. During the training process, we followed a semi-supervised segmentation \cite{AugSeg,advsemiseg2018} data partitioning approach. We sampled 1\%, 5\%, 10\%, 20\%, and 30\% of the training set as labeled data, while the remaining portion was used as unlabeled data. 

\noindent\textbf{Evaluation Metrics.}
Referring to previous work\cite{DGNet,ZoomNet,SegMaR}, we used 6 common evaluation indicators in COD to evaluate our CamoTeacher, including S-measure($S_{\alpha}$) \cite{S_measure} , weight F-measure($F_{\beta}^w$) \cite{wF_measure} , mean E-measure($E_{\phi}^m$) \cite{meanE} , max E-measure($E_{\phi}^x$), mean F-measure($F_{\beta}^m$) \cite{meanF} , and mean absolute error ($M$).
    
\noindent\textbf{Implementation Details.}
%
%
The proposed CamoTeacher is implemented with PyTorch.
We adopt SINet\cite{COD10K} as the baseline of COD model.
We use an SGD optimizer with a momentum of 0.9 and a polynomial learning-rate decay with an initial value of 0.01 to train the student model.
The training epoch is set to 40 with 10 epoch for burn-in phase. And the batch size is 20, with a ratio of 1:1 between labeled and unlabeled data, \ie each batch consists of 10 labeled and 10 unlabeled images. During training and inference, each image is resized to $ 352 \times 352 $.
The teacher model is updated via EMA with a momentum $\eta$ of 0.996.
Weak data augmentation involves Random Flip and Random Scale, while strong data augmentation entails color space transformations including Identity, Autocontrast, Equalize, Gaussian blur, Contrast, Sharpness, Color, Brightness, Hue, Posterize, Solarize, with a random selection of up to 3 from this list.

\subsection{Main Results}
\noindent\textbf{Comparison with Baselines.}
In \cref{tab:result}, we compared CamoTeacher, the supervised baseline \cite{COD10K}, and Mean Teacher \cite{mean-teacher}.
Mean Teacher does not show improvements over the supervised in the semi-supervised setting due to  treating all pseudo-labels and regions equally, which does not account for pseudo-labels noise problem. 
For instance, on the NC4K test set, under the 1\% labeled data, the $S_{\alpha}$ dropped by 9.7\% compared to supervised.
Compared to Mean Teacher, CamoTeacher integrated with DRCL effectively mitigates pseudo-label noise issues and efficiently utilizes pseudo-label information. 
It demonstrates substantial improvements over both supervised and Mean Teacher across various proportions of labeled data, especially in scenarios with extremely limited.
For instance, on 1\% NC4K, CamoTeacher exhibited a substantial increase of 28.6\% in the $F_{\beta}^w$ and a 12.3\% enhancement of $S_{\alpha}$.
Compared with Mean Teacher, the performance gains by CamoTeacher are more significant, which are 38.6\% and 17.7\%, espectively.

In \cref{fig:1}(b), we visualize the performance changes of supervised, Mean teacher and our CamoTeacher on COD10K under different labeled data ratios of 1\%, 5\%, 10\%, and further 20\%, 30\%, which intuitively reflects the effectiveness of CamoTeacher.
It's evident that metrics like $F_{\beta}^w$, $F_{\beta}^m$, $E_{\phi}^m$, and $M$ surpass 100\% of fully supervised performance with just 5\% labeled data. Additionally, with 30\% labeled data, all metrics significantly exceed 100\% of fully supervised performance. 

\noindent\textbf{Application to Existing COD Methods.}
We applied the CamoTeacher approach to additional COD models, including CNN-based model SINet-v2 \cite{SINetv2} and SegMaR \cite{SegMaR} and Transformed-based model DTINet \cite{DTINet} and FSPNet \cite{FSPNet}, and the corresponding results are presented in \cref{tab:diff_COD_model}. In terms of SINet-v2, CamoTeacher demonstrated a 2.8\% improvement in the $S_{\alpha}$ metric and a 6.1\% improvement in the $F_{\beta}^w$ metric compared to the supervised baseline. Based on FSPNet, CamoTeacher achieved a 2.8\% enhancement in the $S_{\alpha}$ metric and a 6.2\% improvement in the $F_{\beta}^w$ metric without utilizing additional discriminative annotation. These outcomes provide robust evidence of the effectiveness of CamoTeacher.

\noindent\textbf{Addition unlabeled data.}
In \cref{tab:addition_data}, based on the base model SINet, we train CamoTeacher with 100\% labeled data and check whether additional unlabeled data can further improve the performance. Compared with unsupervised and weakly-supervised methods, it shows the superiority of our method.
We use the testset CAMO and NC4K as unlabeled data, and we find that adding more additional data CamoTeacher can make SINet continued to gain, such as $S_{\alpha}$ metric increased by 1.6\% and $F_{\beta}^w$ metric increased by 2.5\%.
We achieved significantly better performance than unsupervised methods\cite{SelfMask, FOUND, UCOS-DA} using existing labeled data. 
Compared with the weakly-supervision methods \cite{SCWSSOD, SCOD, WS-SAM}, we do not need additional annotation for the new unlabeled data, and to a certain extent, as the more unlabeled data, the performance will get better and better. And at the same time, replacing the base model with a stronger model will get better results.

\subsection{Ablation Study}
\noindent\textbf{Ablation on PCL and ICL.}
We conducted ablation experiments on different combinations of weight and loss functions for PCL and ICL. From \cref{tab:ablation_mian}, increasing the weights $\omega^{pc}$ and $\omega^{ic}$ improves the model's performance. Specifically, with $\omega^{pc}$ resultes in a higher $F_{\beta}^w$ score, increasing from 0.370 to 0.501. Similarly, with $\omega^{ic}$ improved the $F_{\beta}^w$ metric from 0.572 to 0.585. Combining these two weighted loss functions led to a significant performance improvement, with the $F_{\beta}^w$ score increasing from 0.370 to 0.586 (an increase of 21.6\%). This demonstrates the effectiveness of PCL and ICL in enhancing the model's performance.

\noindent\textbf{Ablation of weight approach in PCL.}
The choice of method for obtaining $\omega^{pc}$ significantly affects the performance of the model. Therefore, we conducted several experiments to select the most effective method for obtaining $\omega^{pc}$. From \cref{tab:weight}, it can be observed that the best performance is achieved when $\omega^{pc}$ is obtained using \cref{wlc}.

\noindent\textbf{Ablation of hyper-parameters in PCL and ICL.}
We perform ablation experiments on the hyper-parameters associated with the PCL and ICL modules. As shown in \cref{tab:hyper_alpha}, the optimal results are achieved when setting the hyper-parameter $\alpha$, which corresponds to the coefficient $\omega^{pc}$, to 1/4.
Similarly, according to \cref{tab:hyper_beta}, the best performance is observed when the hyper-parameter $\beta$, related to the coefficient $\omega^{ic}$, is set to 4.


\subsection{Qualitative Analysis}
\noindent\textbf{Result visualization.}
We conducte visualizations of CamoTeacher's results under varying proportions of labeled data. The model is trained with 1\%, 5\%, 10\%, 20\% and 30\% labeled data, and we compare its predicted masks with those produced by SINet under fully supervised conditions (as shown in \cref{fig:vis_main}). Notably, even with only 10\% labeled data, CamoTeacher's predicted masks exhibit a remarkable similarity to the fully supervised SINet predictions. In fact, in certain detailed predictions, CamoTeacher even surpasses the performance of fully supervised SINet. These findings highlight the impressive capability of CamoTeacher in leveraging limited labeled data for effective camouflaged object detection.

\section{Conclusions}
\label{conclusions}
%
In this paper, we present the first end-to-end semi-supervised camouflaged object detection model called CamoTeacher.
To address the issue of substantial noise within pseudo-labels in semi-supervised COD, including both local noise and global noise, we've introduced a novel approach called Dual-Rotation Consistency Learning (DRCL), including Pixel-wise Consistency Learning (PCL) and Instance-wise Consistency Learning (ICL). 
DRCL assists the model in alleviating noise issues, effectively leveraging pseudo-labels information, so that the model can receive adequate supervision while avoiding confirmation bias.
Extensive experiments validate CamoTeacher's superior performance while significantly reducing annotation expenses.
We believe that CamoTeacher can serve as a strong baseline for future research on semi-supervised COD.



\section*{Acknowledgements}
\label{acknowledgements}
This work was supported by National Science and Technology Major Project (No. 2022ZD0118201), the National Science Fund for Distinguished Young Scholars (No.62025603), the National Natural Science Foundation of Chine a (No. U21B2037, No. U22B2051, No. U23A20383, No. 62176222, No. 62176223, No. 62176226, No. 62072386, No. 62072387, No. 62072389, No. 62002305 and No. 62272401), and the Natural Science Foundation of Fujian Province of China (No.2022J06001).

%
\bibliographystyle{splncs04}
\bibliography{main}
\end{document}